\documentclass[graybox]{svmult}

\usepackage{type1cm}        
                            
\usepackage{makeidx}         
\usepackage{graphicx}        
                             
\usepackage{multicol}        
\usepackage[bottom]{footmisc}

\usepackage{newtxtext}        
\usepackage[varvw]{newtxmath}

\makeindex

\begin{document}

\title*{\small \textit{This is a preprint of a chapter accepted for publication in Generative and Agentic AI Reliability:
Architectures, Challenges, and Trust for Autonomous Systems, published by Springer Nature.} \\ [0.5cm]
       \Huge \textbf{Perspectives on a Reliability Monitoring Framework for Agentic AI Systems}}

\author{Niclas Flehmig\orcidID{0009-0006-6458-2689}, Mary Ann Lundteigen\orcidID{0000-0002-9045-6815} and\\ Shen Yin\orcidID{0000-0002-3802-9269}}

\institute{Niclas Flehmig \at Norwegian University of Science and Technology (NTNU), Department of Mechanical and Industrial Engineering, Trondheim, Norway, \email{niclas.flehmig@ntnu.no}
\and Mary Ann Lundteigen \at NTNU, Department of Engineering Cybernetics, Trondheim, Norway
\and Shen Yin \at NTNU, Department of Mechanical and Industrial Engineering, Trondheim, Norway}

\maketitle

\abstract*{The implementation of agentic AI systems has the potential of providing more helpful AI systems in a variety of applications. These systems work autonomously towards a defined goal with reduced external control. Despite their potential, one of their flaws is the insufficient reliability which makes them especially unsuitable for high-risk domains such as healthcare or process industry. Unreliable systems pose a risk in terms of unexpected behavior during operation and mitigation techniques are needed. In this work, we derive the main reliability challenges of agentic AI systems during operation based on their characteristics. We draw the connection to traditional AI systems and formulate a fundamental reliability challenge during operation which is inherent to traditional and agentic AI systems. As our main contribution, we propose a two-layered reliability monitoring framework for agentic AI systems which consists of a out-of-distribution detection layer for novel inputs and AI transparency layer to reveal internal operations. This two-layered monitoring approach gives a human operator the decision support which is needed to decide whether an output is potential unreliable or not and intervene. This framework provides a foundation for developing mitigation techniques to reduce risk stemming from uncertain reliability during operation.}

\abstract{The implementation of agentic AI systems has the potential of providing more helpful AI systems in a variety of applications. These systems work autonomously towards a defined goal with reduced external control. Despite their potential, one of their flaws is the insufficient reliability which makes them especially unsuitable for high-risk domains such as healthcare or process industry. Unreliable systems pose a risk in terms of unexpected behavior during operation and mitigation techniques are needed. In this work, we derive the main reliability challenges of agentic AI systems during operation based on their characteristics. We draw the connection to traditional AI systems and formulate a fundamental reliability challenge during operation which is inherent to traditional and agentic AI systems. As our main contribution, we propose a two-layered reliability monitoring framework for agentic AI systems which consists of a out-of-distribution detection layer for novel inputs and AI transparency layer to reveal internal operations. This two-layered monitoring approach gives a human operator the decision support which is needed to decide whether an output is potential unreliable or not and intervene. This framework provides a foundation for developing mitigation techniques to reduce risk stemming from uncertain reliability during operation.}

\section{Introduction}
\label{sec:1}

The emergence of agentic AI systems, systems that adaptively pursue complex goals with limited direct oversight \cite{b1}, has achieved significant attention. Many sectors may benefit from taking such systems into use, such as in healthcare, education, and the process industry. In healthcare, they could enhance diagnostic accuracy and personalize therapies, in education, they could provide intelligent tutoring to socially and economically marginalized groups and in industrial settings, they could reduce cognitive load by offering situational-aware decision support. This potential extends to safety-critical systems, promising more responsive and helpful AI systems.\\
For instance, consider an agentic AI system within a hydrogen bunkering process, responsible for maintaining safety in terms of liquid hydrogen leakage. Such an agent could continuously monitor sensor data to identify patterns indicative of a liquid hydrogen leak, and autonomously inform an operator and initiate mitigation measures like ventilation. However, the implementation of highly autonomous AI in these contexts raises profound reliability concerns. This is reflected in the technical report ISO/IEC TR 5469 \cite{b2} from the International Standard Organisation (ISO) and the International Electrotechnical Commission (IEC), which, while not explicitly naming agentic AI system addresses highly autonomous systems. It suggests that such systems likely fall into a risk category where current methods are insufficient to adequately mitigate reliability-related risks.\\
To mitigate the risk arising from insufficient reliability performance of agentic AI systems, an appropriate monitoring is essential to detect and enable human intervention in critical situations. This requirement for monitoring is underscored by regulations like the EU AI Act \cite{b3}, which mandates post-market monitoring for high-risk AI systems, a category that would include agentic AI systems in safety-critical applications such as hydrogen bunkering. Although the EU AI Act specifies the requirement for appropriate monitoring rather than prescribing exact techniques, reliability monitoring is widely recognized as a part of a monitoring strategy to ensure that AI systems satisfy their functional requirements during operation \cite{b4}.\\
The monitoring of AI systems gained significant relevance as applications moved beyond research environments into real-world deployments \cite{b5, b6}. Initially, monitoring objectives were closely aligned with quality assurance, focusing on continuous performance tracking and improvement. However, early methods that proved effective on curated test sets often failed during operational deployment \cite{b5}. This persistent gap stems from the unstructured nature of real-world inputs and the absence of ground truth labels, which are inherent to production environments \cite{b5, b6}.\\
This insufficiency of early monitoring techniques has driven considerable innovation in recent years, leading to an expansion of the field's objectives. Contemporary research \cite{b4, b7, b8} reflects a paradigm shift from merely detecting incorrect outputs at the component level to ensuring holistic AI safety at the system level. Consequently, monitoring approaches can be classified into distinct types based on their specific objectives \cite{b4}. Current monitoring tools for AI systems \cite{b9, b10, b11, b12} do not provide a comprehensive framework specifically designed for monitoring the operational reliability of agentic AI systems, as their objectives are often broader and they are applied to non-agentic AI systems.\\
In this work, we derive the reliability challenges of agentic AI systems from their characteristics which distinguish them from traditional AI systems. Moreover, we compare these challenges to the reliability challenges of traditional AI systems and formulate a common fundamental challenge to both. Based on this, we propose a two-layered reliability monitoring framework for agentic AI systems that identifies novel inputs to the system, and reveals internal operations of the agentic AI system to enable a human operator to make a decision whether a current state of the system is unreliable or not and trigger a fallback policy. This monitoring framework can be seen as mitigating technique for risk stemming from insufficient reliability performance.
\section{Definitions}
\label{sec:2}
\subsection{Agentic AI systems}
\label{sec:21}
Agentic AI systems are defined as AI systems capable of autonomously performing actions to achieve a specified goal, without relying on pre-defined behavioral scripts \cite{b1}. Unlike traditional AI systems like ChatGPT that execute specific, bounded tasks, agentic AI systems are characterized by a higher degree of agenticness. This degree is determined by four key characteristics: goal complexity, environmental complexity, adaptability, and autonomy \cite{b1, b13}.\\
For instance, ChatGPT alone would not be considered agentic \cite{b1}. Instead, an agentic AI system could autonomously conduct an entire literature review: searching for papers, filtering results, and summarizing findings based on a high-level research query provide by a human \cite{b13}. This illustrates the shift from reactive task-completion to proactive goal-pursuit.\\
This work focuses specifically on agentic AI systems that operate under human supervision, pursuing human-defined goals within human-defined environments as collaborative partners.
\subsection{Reliability of AI systems}
\label{sec:22}
The term \textit{reliability} has no standardized definition and varies from industry to industry. One possible definition for technical systems describes an entity's capability to deliver required performance over a defined period and under specified operational conditions \cite{b14}. For AI systems including agentic AI systems, the reliability can be defined as assessment of the model's performance consistency over time and under varying conditions \cite{b7}. While the definition of reliability does not differ significantly, the definition of the term \textit{performance} does. For technical systems, the performance can be determined by laws and regulations, standards, and customer requirements \cite{b14}. For AI systems, the performance is determined by executing a set of tasks relevant to the desired functionality of the system. Such sets of tasks are sometimes standardized under so-called benchmarks \cite{b15}. In general, there is no standardized performance metrics for AI systems.
\section{Methods}
\label{sec:3}
\subsection{Research approach} \label{sec:31}
In this work, we adopted a qualitative, conceptual research design. The primary objective is to address a complex, emerging challenge which is to monitor the operational reliability of agentic AI systems. The research was conducted in two sequential phases:
\begin{enumerate}
    \item \textbf{Identification of the main reliability challenge for operational agentic AI systems}: A comprehensive literature review was undertaken to investigate the operational reliability challenges of agentic AI systems based on their characteristics. These challenges were compared with those of traditional AI systems to identify a common fundamental challenge.
    \item \textbf{Development of reliability monitoring framework for agentic AI systems}: Building upon the identification of the challenge, a conceptual two-layered framework was constructed. This involved a second literature review to identify relevant monitoring techniques from traditional AI systems and integrate them into a cohesive structure for detection and decision-support.
\end{enumerate}
Given the novelty of the problem, a traditional systematic literature review protocol was deemed less suitable. Instead, an exploratory, iterative literature review methodology was employed.
\subsection{Data collection and analysis}\label{sec:32}
The literature review was rather iterative than linear, characterized by a snowballing technique and conceptual branching. The three core topics were:
\begin{itemize}
    \item \textbf{Characteristics of agentic AI systems}: The review began with literature on \textit{agentic AI systems} \cite{b1, b13, b16, b17, b18, b19, b20}. The focus here was to understand the typical characteristics of agentic AI systems, their applications, and their technical challenges to derive how these affect the operational reliability. This phase was important for ensuring the relevance of later findings to the agentic AI systems context.
    \item \textbf{Reliability challenges of traditional AI systems}: From the characteristics of agentic AI systems, the review progressed into reliability challenges in traditional AI systems. For this, literature on \textit{AI deployment} \cite{b6, b21} and \textit{AI monitoring} \cite{b4, b5, b8, b9, b22} was conducted to establish an understanding, draw connections to the agentic AI systems, and identify similarities.
    \item \textbf{Focused exploration on AI monitoring concepts, frameworks and techniques}: Finally, the review process concluded into more specific areas based on references and linked concepts within the initial literature. The key threads that were followed included:
    \begin{itemize}
        \item Broader concept of AI monitoring as part of AI safety \cite{b7, b23, b24}
        \item Conceptual frameworks and approaches for monitoring AI systems \cite{b9, b12, b22, b25}
        \item Techniques for out-of-distribution (OOD) detection in model inputs \cite{b26, b27, b28, b29}
        \item Techniques for AI transparency and explainable AI (XAI) \cite{b7, b30, b31} 
    \end{itemize}
    This phase of the review enabled a synthesis of existing monitoring frameworks and techniques, providing the foundation for proposing a new approach for agentic AI systems.
\end{itemize}
The main sources for the literature review were \textit{Web of Science}, \textit{Google Scholar}, and \textit{arXiv}.
\subsection{Underlying assumptions}\label{sec:33}
This work is based on the following assumptions regarding pre-deployment lifecycle stages and operational context of the agentic AI systems:
\begin{itemize}
    \item \textbf{Pre-deployment development}: All prior lifecycle stages including data acquisition, model selection, training, testing, and verification were conducted optimally. The system achieved a perfect reliability score on an adequate dataset, meaning the data is balanced and free of bias. This is, in practice, impossible to achieve for complex agentic AI systems but it isolates the challenges that emerge specifically during operation.
    \item \textbf{Deployment environment}: The systems are deployed in environments appropriate to their agentic nature, characterized by complexity and unpredictable dynamics as described in the literature \cite{b1, b16}.
    \item \textbf{Scope of operational reliability for agentic AI systems}: The scope of operational reliability is limited to failures arising from the agentic properties of the system. We explicitly assume a reliable underlying infrastructure, thereby excluding issues related to software dependencies, API failures, or hardware faults. Moreover, the system operates within a secured environment. We assume that potential threats from adversarial inputs are mitigated by appropriate cybersecurity measures and thus, this is outside the scope of this work.
    \item \textbf{Performance metric for reliability assessment}: Currently, there is no standardized performance metric for agentic AI systems \cite{b16}. Therefore, we adopt the task success rate as metric for assessing performance and thereby reliability. This metric is defined as proportion of correctly executed tasks required to achieve a specified goal.
\end{itemize}
By holding these factors constant, this chapter can focus specifically on the monitoring of reliability during operation for agentic AI systems.
\section{Results}\label{sec:4}
This section presents the two core outcomes of this work: (1) the identification of an unpredictable environment as main reliability challenge for agentic AI systems and traditional AI systems during operation, and (2) the proposed two-layered reliability monitoring framework for agentic AI systems.
\subsection{The reliability challenge for agentic AI systems}\label{sec:41}
\subsubsection{Characteristics of agentic AI systems and their reliability implications}\label{sec:411}
\begin{figure}[b]
    \centering
    \includegraphics[width=0.75\textwidth]{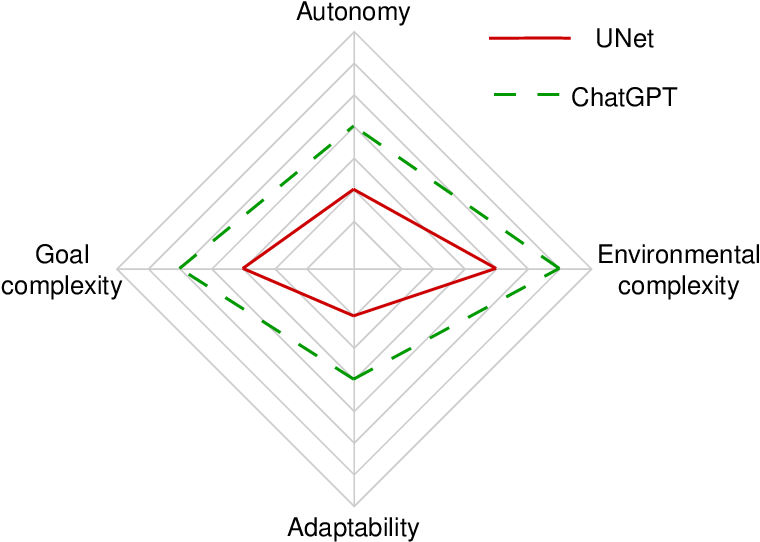}
    \caption{Exemplary generic radar chart for the qualitative assessment of the agenticness of an AI system. The solid line represents a UNet model for image segmentation and the dashed line represents a ChatGPT model.}
    \label{fig:AgenticAIRadar}       
\end{figure}

Based on literature \cite{b1, b16, b20, b32}, the characteristics for agentic AI systems can be grouped into four key areas: \textit{Autonomy}, \textit{Goal complexity}, \textit{Environmental complexity}, and \textit{Adaptability}. These can be also seen as the characteristics that differentiate them from traditional AI systems. Using a radar chart, Fig. \ref{fig:AgenticAIRadar} shows how a qualitative assessment of the agenticness of an AI system could look like where the degree of each characteristic increases with the distance to the center. For the purpose of identifying impacts on reliability during operation which meant to be monitored to ensure reliability, we must assess which of these characteristics directly impact the reliability of agentic AI systems during operation.
\\
\\
\begin{itemize}
    \item \textbf{Autonomy}: Autonomy refers to the capability of agentic AI systems to execute a sequence of tasks and make strategic decisions with minimal or no human intervention to achieve a goal \cite{b16, b32}. From our reliability perspective, where performance is measured by successful completion of defined tasks, this autonomy may not directly impact reliability. However, it introduces a significant risk of value misalignment, where the system's chosen method for achieving a goal conflicts with human values, which can be perceived as a form of unreliability from a broader safety perspective \cite{b7}.
    \item \textbf{Goal complexity}: Goal complexity refers to an agentic AI system's capacity to achieve an objective by executing a wide range of distinct tasks. For example, unlike a traditional image classifier, an agentic system could analyze an image, edit it based on user input, and generate a descriptive caption \cite{b1}. While achieving such complex goals requires a powerful and well-designed system, this complexity is    primarily a pre-deployment consideration. The inherent complexity of a goal does not, in itself, directly impact operational reliability, as the system's capability to handle the variety of tasks is established before deployment.
    \item \textbf{Environmental complexity}: Environmental complexity refers to the dynamic nature of the environments where agentic AI systems are deployed. These environments can be cross-domain, involve multiple stakeholders, and require the use of external tools, leading to a high degree of unpredictability. For example, while a traditional AI system might master the fixed rules of chess, an agentic AI system must adapt to the vast and variable rules of any board game \cite{b1, b16}. This characteristic has a direct and significant impact on operational reliability. Pre-deployment testing can only represent a limited snapshot of potential real-world scenarios. As the environment evolves with unpredictable dynamics, the system may encounter novel situations that challenge its trained capabilities, directly affecting its reliability during operation \cite{b16}.
    \item \textbf{Adaptability}: Adaptability is the capacity of agentic AI systems to respond to novel and unexpected circumstances, typically enabled by techniques like reinforcement learning and meta-learning \cite{b1, b16, b32}. The relationship between adaptability and operational reliability is complex. In theory, adaptability is intended to enhance reliability by allowing the system to handle environmental dynamics. However, the current insufficiency of the commonly implemented methods become a direct source of operational unreliability. When a system fails to adapt adequately to unforeseen situations, its reliability is compromised \cite{b16}.
\end{itemize}
In summary, the operational reliability of agentic AI systems is most significantly impacted by the interplay between environmental complexity and adaptability. The unpredictable nature of real-world environments makes them impossible to fully anticipate before deployment, while current adaptive capabilities are insufficient to respond to this complexity effectively. Consequently, the gap between a unpredictable environment and a system's limited adaptability directly undermines reliability during operation.

\subsubsection{Comparison of reliability challenges between traditional and agentic AI systems}\label{sec:412}
While traditional AI systems lack the agenticness which is defined by the characteristics mentioned above, their operational reliability is impacted by the same fundamental issue: environmental unpredictability. For traditional systems, this manifests primarily through data drift and outliers \cite{b6, b21}.\\
Data drift, defined as a shift in the underlying data distribution \cite{b26, b33, b34}, occurs due to unpredictable environmental changes. This phenomena can be categorized into two main types \cite{b26}:
\begin{itemize}
    \item \textbf{Semantic shift}: This affects the decision boundary of a model. It results in a shift of the label space which implies the alteration of existing labels or even introduce new ones.
    \item \textbf{Covariate shift}: This shift only affects the input space, while the label space remains constant. Adversarial examples \cite{b35}, or domain shift \cite{b36} are examples for such a covariate shift.
\end{itemize}
The distinction between these shifts is critical for reliability. Covariate-shifted data serves as a valuable test for an AI system's generalization and robustness \cite{b26}. Generalization examines how well an AI system can perform on unseen data and robustness describes the AI system's property to behave safely in a wide range of circumstances \cite{b23}. A well-trained system can maintain reliability against such a shift. Semantic shift, however, poses a more severe threat to operational reliability, as it invalidates the core rules the system learned \cite{b26}. This is where agentic AI systems potentially hold an advantage with their capability for adaption in complex environments \cite{b1, b16, b32}. This may allow them to handle semantic shifts more effectively than traditional systems, which require retraining \cite{b6, b33, b37}. Nonetheless, the root cause which is the environmental unpredictability, remains a challenge for traditional AI systems and for agentic AI systems, despite potentially different degrees.\\
The second mentioned challenge, outliers, can be viewed as an instance-level analog to data drift. Instead of a distribution shift over time, an outlier is a single instance that deviates from the trained distribution \cite{b26}. Although not always explicitly discussed in the context of agentic AI systems, outliers are an inherent part of environmental complexity and represent inputs outside a system's learned experience.\\
Both data drift and outliers can be unified under the broader concept of \textit{out-of-distribution} (OOD) data \cite{b26}. Here, an outlier is a single OOD instance, while data drift represents a sequence of OOD instances. This unified framework simplifies the analysis of operational reliability impacts.\\
In conclusion, despite their different characteristics, traditional and agentic AI systems face a common challenge to operational reliability: unpredictable environmental dynamics that produce OOD data. This fundamental similarity suggests that techniques developed for monitoring OOD data in traditional AI systems remain highly relevant for ensuring the reliable operation of agentic AI systems.

\subsection{Reliability monitoring framework for agentic AI systems}\label{sec:42}
As established in Section \ref{sec:31}, the fundamental challenge to operational reliability is the unpredictable environment, which produces OOD instances that exposes the limitations of the agentic AI system's adaptability. This section proposes a reliability monitoring framework designed to mitigate this challenge through a two-layered approach: detection followed by decision-support. The first layer employs an appropriate OOD detection technique to identify and flag novel inputs. However, as OOD alerts alone are not sufficient in determining the consequence of novel inputs \cite{b9, b26}, the second layer incorporates AI transparency to contextualize these alerts. This integration enables a distinction between a failure mode and a system's successful adaptation to new circumstances.

\subsubsection{First layer: out-of-distribution detection as environmental sensor}\label{sec:421}
Out-of-distribution (OOD) detection acts as a sensor for the environment, tasked with evaluating whether a given input deviates from the agentic AI system's learned data distribution. Formally, this is the problem of distinguishing in-distribution (ID) data from OOD data \cite{b26}. A critical prerequisite for effective OOD detection is a clear separation between the learned distribution and OOD instances because an overlap makes the distinction unlearnable \cite{b28}.\\
To clarify the data landscape, we adapt a taxonomy that identifies three key categories \cite{b29}:
\begin{itemize}
    \item \textbf{In-distribution (ID) data}: Data that aligns with the training distribution (learned distribution).
    \item \textbf{Covariate-shifted ID data}: Data where the input distribution has shifted, but the underlying output distribution remains valid.
    \item \textbf{OOD data}: Data that represents a semantic shift, where the output distribution is shifted as well.
\end{itemize}
In theory, OOD detection encompasses identifying both covariate-shifted ID data and semantic-shifted OOD data, as both lie outside the original training distribution. However, the prevailing research focus is on detecting semantic shift, often treating covariate shift as a test for generalization, as mentioned in Section \ref{sec:412}, rather than a target for detection. Including covariate shift as a detection target is controversial, as it can lead to a high rate of false positives, flagging inputs that a robust system should handle correctly \cite{b26}.\\
The choice of a practical OOD detection technique often depends on data availability. A common scenario is having only ID data available for training the detector. Reconstruction-based methods are well-suited for this. For example, the \textit{LMD} (Lift, Map, Detect) technique \cite{b38} uses a diffusion model to learn a compressed representation of the ID data manifold. It detects OOD instances by masking part of an input image, lifting it off the manifold, using the model to inpaint the missing part, mapping it back, and then measuring the reconstruction error. A small distance between the original image and the inpainted image indicates ID data. An agentic AI system processing visual inputs could leverage such a method to flag unfamiliar environments.\\
In contrast, zero-shot methods require no model training on ID data. Approaches like \textit{Concept-based Zero-shot OOD Detection} \cite{b39} utilize large pre-trained vision-language models like CLIP \cite{b40}. They calculate an ID score based on a positive concept derived by using LLMs, a negative concept computed by using NegMining \cite{b41}, and a base label set. This score determines the OOD classification. Such methods are advantageous as they avoid extensive training of monitoring models.\\
The examples above represent a small fraction of the emerging field of OOD detection techniques, which offers a wide range of techniques for different domains \cite{b26, b27}. However, these methods face common challenges. First, many techniques are validated empirically rather than with formal guarantees \cite{b28}, making it difficult to confidently transfer them from benchmarks to real-world applications. Second, a significant number of detectors are more sensitive to covariate shift than to semantic shift \cite{b9, b29}. This can result in a high rate of false positives.\\
Especially, false positives reveal a limitation for our framework where an OOD alert signals novelty but not necessarily a failure mode. Agentic AI systems are designed to adapt. A detector prone to false positives would be impractical. Therefore, the binary signal from OOD detection is insufficient on its own. It must be complemented with additional information to understand the behavioral impact of the novelty and determine if a true reliability failure is occurring.
\subsubsection{Second layer: AI transparency as decision-support}\label{sec:422}
Since OOD detection alone cannot determine whether a novel input will lead to an unreliable output, a decision-support is required to assess the agent's internal response. AI transparency serves as this essential second layer, aiming to reveal a model's internal operations and provide understandable explanations for the causal connection between its inputs and outputs \cite{b7, b30}.\\
It is crucial to distinguish this from simply revealing a model's parameters. The immense scale of modern networks, such as DeepSeek-V3 with 671 billion parameters \cite{b42}, renders raw parameter data uninterpretable. This challenge is magnified in agentic AI systems, which often function as advanced, multi-step LLMs \cite{b1, b13, b16}. Therefore, practical transparency relies on specialized techniques to summarize and interpret complex behaviors. These techniques can be broadly categorized into three main approaches: \textit{Explainability}, \textit{Mechanistic interpretability}, and \textit{Representation engineering} \cite{b7}.\\
\textit{Explainability} refers to a class of algorithms that explicitly incorporate human comprehensibility as an objective when generating justifications for a model's decisions \cite{b43}.\\
A visual technique for explainability is saliency maps, first developed for image data, that highlights pixels that were most relevant for a model's classification decision \cite{b44}. The intuition is based on human behavior, mimicking human focus on salient object features. Despite producing visually compelling heatmaps, a limitation is their lack of actual information about the model's actual internal operations. Studies have shown, they often function as mere bias-confirming tools, producing similar outputs for both trained and randomly initialized models, thus failing to provide genuine insight \cite{b45}.\\
For LLMs, \textit{Chain-of-Thought} (CoT) prompting has emerged as a technique for producing explanations \cite{b46}. CoT aims to make the model's "thinking" visible by generating a step-by-step reasoning process before the final answer. However, a major concern is the faithfulness of these generated "thoughts". The CoT output is itself a generation of intermediate steps of the LLM which might be filtered before printed. It may constitute plausible sounding reasoning but it does not actually reflect the true casual path the model took to arrive at the answer which may be potentially misleading \cite{b47,b48}.\\
In contrast to explainability's post-hoc nature, \textit{mechanistic interpretability} is a bottom-up approach that seeks a complete, causal understanding of a model's internal computations. It aims to reverse-engineer neural networks into human-understandable algorithms and circuits \cite{b49}.\\
The methodology centers on identifying features, directions within a model's activation space that correspond to coherent, meaningful concepts \cite{b50}. For instance, an LLM might develop a feature for "this is Paris". When processing the input "Eiffel Tower", the activation vector would align with this feature, whereas "Big Ben" would not \cite{b54}. The goal is to find a universal set of such mono-semantic features.\\
Sparse autoencoders have been successfully used to decompose LLM activations into dictionaries of sparse, potentially interpretable features \cite{b51}. However, current methods show limitations, including ambiguity in feature semantics and a failure to consistently produce a truly universal and mono-semantic feature set \cite{b52}.\\
\textit{Representation engineering}  offers a top-down alternative to mechanistic interpretability's bottom-up approach \cite{b53}. It operates on the premise that a model's internal representations encode meaningful concepts, even if they are not directly human-interpretable. Instead of fully reverse-engineering the model, it develops tools to directly read and control these representations.\\
The core technique involves using contrastive concepts. By prompting a model with contrasting pairs of inputs, e.g. reliable and unreliable input, and measuring the differential activation patterns, it becomes possible to isolate the model's "steering vectors" for specific behaviors. These vectors can then serve as detectable signals for identifying potentially unreliable outputs \cite{b7}.\\
Despite progress, transparency approaches have some universal challenges. First, many lack robustness and consistency, performing well on benchmarks but failing under deeper analysis, as seen in saliency maps and sparse autoencoder \cite{b45, b52}. Second, they often do not scale effectively to the complexity of modern generative models and thus agentic AI systems \cite{b30}. Most critically, there is a human-centricity gap. These approaches are typically designed by experts for experts, neglecting the needs of end-users who must interpret their outputs. If those assessing reliability cannot understand the explanations provided, these tools risk being misleading rather than enlightening \cite{b30, b31}.
In the context of our framework, these transparency techniques provide the critical context needed to decide whether the alerts raised by OOD detection cause an unreliable output or not, moving from simply detecting novelty to understanding its behavioral impact and making a decision.
\subsubsection{Application of the reliability monitoring framework}\label{sec:423}
To ensure the operational reliability of agentic AI systems, we propose a two-layered monitoring designed to detect and support the decision on potentially unreliable outputs. This framework addresses the core challenge of unpredictable environments by first identifying novel inputs that deviate from the system's learned experience, then reveals the internal operations of the agentic AI systems and thereby supporting the decision on unreliable outputs. The following explains the framework's pipeline and concludes with a simple illustrative example.\\
The monitoring process begins with the first layer: \textit{OOD detection}. This layer continuously observes the data streams entering the agentic AI system. Since agents may process multimodal inputs, like images, audio, text, a separate OOD detector should be deployed for each data source.  When any detector identifies an OOD instance, it triggers the second layer of the framework.\\
The second layer: \textit{AI transparency}, is then activated. This component aims to reveal the internal operations of the agentic AI systems in response to the flagged input. Unlike the first layer, a single transparency monitor for the entire agent is typically sufficient. However, its analysis must be contextualized by the specific data stream that triggered the OOD alert to generate a relevant explanation of the system's behavior, such as its reasoning trace or decision-making process.\\
In the final step, a human operator reviews the synthesized information from both layers: the OOD alert indicating what is novel and the transparency report explaining how the agent is handling it. This combined insight enables an informed judgment on whether the novel input is leading to an unreliable output or if the agent is adapting correctly. The operator can then decide on an appropriate action, such as interrupting the agent, allowing it to proceed, or initiating a conservative fallback policy.\\
For instance, consider again the agentic AI system responsible for safety during a hydrogen bunkering process. Its goal is to prevent liquid hydrogen leaks by analyzing real-time sensor data and initiate appropriate control actions. An OOD detector, monitoring the pressure sensor stream, flags a novel input not present in the training data. This triggers the transparency monitor, which captures the agent's internal reasoning trace via CoT. The human operator sees that while the sensor input is novel (OOD), the agent has correctly identified a potential precursor to a leak and has initiated a conservative control action. The operator can therefore acknowledge the OOD alert not as a failure mode, but as a successful detection of a novel scenario that the agent is handling appropriately by adapting.\\
In summary, this framework moves beyond just OOD detection by integrating decision-supporting transparency, thereby providing a practical mechanism for managing the operational reliability of agentic AI systems.
\section{Discussion}\label{sec:5}
This work explores the critical challenge of maintaining operational reliability in agentic AI systems. While these systems possess distinct characteristics that differentiate them from traditional AI systems \cite{b1, b16, b32}, they nevertheless face a fundamental and shared challenge: an unpredictable environment. To address this, we proposed a novel two-layered reliability monitoring framework designed as a practical guide for developing such operational reliability monitoring tools.
\subsection{Reliability challenge of operational agentic AI systems}\label{sec:51}
The findings of this work show that the fundamental challenge to operational reliability, which is the unpredictable environment, is conceptually consistent across both traditional and agentic AI systems. We acknowledge that the definition of reliability is inherently tied to the selection of performance metrics, which influences the identification of potential challenges. Incorporating additional metrics, such as resource efficiency and long-term goal achievement \cite{b16}, would increase the complexity of reliability monitoring and impact the challenges. However, the absence of standardized performance evaluation for agentic AI systems \cite{b32}, coupled with critiques against combining numerous metrics \cite{b15}, led us to solely adopt task success rate for its clarity and straightforward interpretability.\\
Reducing the operational reliability challenge to the core concept of OOD detection could be viewed as oversimplification, particularly as it may not fully capture temporal aspects like concept drift \cite{b33}. Nevertheless, this abstraction serves to isolate the central problem, which is that the system encounters unseen inputs. This simplification provides a foundational and manageable lens through which to address the challenge.\\
By establishing the connection between traditional and agentic AI systems through the common reliability challenge, our work highlights an opportunity to adapt and transfer knowledge and monitoring techniques from traditional AI systems \cite{b5, b6, b8, b9, b22}. This bridges a gap, offering a practical starting point for ensuring their reliable operation.
\subsection{Two-layered reliability monitoring framework}\label{sec:52}
The proposed two-layered monitoring framework offers a structured approach to ensure the operational reliability of agentic AI systems. By integrating OOD detection to flag novel inputs with AI transparency to reveal internal reasoning, it provides a human operator with the contextual data needed to decide on potential reliability failures. This moves beyond a simple binary alert towards a informed decision process.\\
The rationale for this layered approach is grounded in practical limitations. Relying solely on OOD detection has proven problematic, often generating a high rate of false positives that can undermine its utility \cite{b9, b26}. Introducing a second, diagnostic layer addresses this by providing context, an approach aligned with recommended AI safety practices that advocate for combined input monitoring and transparency \cite{b23}. Transparency techniques are particularly vital for making the complex operations of agentic AI systems comprehensible, thereby supporting human judgment \cite{b7, b30}. However, it is critical to acknowledge that these explanations can be misleading if not properly designed and interpreted, underscoring the importance of selecting appropriate tools and training operators effectively \cite{b30, b31}.\\
The framework's reliance on a human-in-the-loop presents both a strength and a limitation. To prevent operator desensitization, a design that fosters continuous interaction during both detecting a novel input and normal operations is preferable to one based solely on emergency interventions. Furthermore, the need for human analysis inherently limits the framework to scenarios where real-time decision-making is not required. Consequently, this approach is best suited for high-stakes domains like healthcare or safety-critical systems in process industry, where the cost of error justifies the investment in human oversight and the decision-making timeline permits deliberation.\\
Finally, we acknowledge that the framework, while grounded in a thorough review of literature and our own expertise, currently lacks extensive empirical validation. Its value at this stage is primarily as a conceptual blueprint and a practical guideline for structuring the development of reliability monitoring tools for agentic AI systems, with future experimental work being a necessary next step.
\section{Conclusion}\label{sec:6}
In summary, this work identifies the core operational reliability challenges for agentic AI systems and establishes their fundamental similarity to those of traditional AI systems. They are both affect by unpredictable environments that generate OOD data. To address this, we have proposed a novel two-layered monitoring framework. The first layer employs OOD detection to flag novel inputs that deviate from the system's learned distribution. The second layer incorporates AI transparency techniques to reveal the systems internal operations.\\
This two-layered approach is important because while OOD detection signals novelty, it cannot determine the response of the agentic AI system. Given that agentic AI systems possess adaptability capabilities and not all OOD instances lead to failure modes, the transparency layer provides the necessary context for decision-making. Together, they empower a human operator to distinguish between a critical reliability failure mode and a successful adaptation.\\
By providing a structured approach to a common challenge, this framework serves as a valuable guide for developing practical monitoring tools to ensure the reliable operation of agentic AI systems in real-world deployments. It may help to develop techniques for reducing risk that stems from highly autonomous systems in safety-critical systems. Future work will focus on the experimental validation of this approach.
\begin{acknowledgement}
This work was carried out as a part of SUBPRO-Zero, a Research Centre at the Norwegian University of Science and Technology. The authors gratefully acknowledge the project support from SUBPRO-Zero, which is financed by major industry partners and NTNU.
\end{acknowledgement}


\begin{thebibliography}{99.}%

\bibitem{b1} Shavit, Yonadav, Agarwal, Sandhini, Brundage, Miles, Adler, Steven, O’Keefe, C., Campbell, R., Teddy Lee, Pamela Mishkin, Tyna Eloundou, Alan Hickey, Katarina Slama, Lama Ahmad, Paul McMillan, Alex Beutel, Alexandre Passos, \& David G. Robinson. (2023). Practices for Governing Agentic AI Systems (Practices for Governing Agentic AI Systems) [White paper]. OpenAI. https://openai.com/index/practices-for-governing-agentic-ai-systems/

\bibitem{b2} Artificial intelligence -Functional safety and AI systems ISO/IEC TR 5469. (n.d.). [Standard]. Retrieved April 5, 2024, from https://www.iso.org/standard/81283.html\#lifecycle

\bibitem{b3} EU AI Act. (2024). European Union Artificial Intelligence Act. European Union. https://artificialintelligenceact.eu/ai-act-explorer/

\bibitem{b4} Yampolskiy, R. V. (2024). On monitorability of AI. AI and Ethics. https://doi.org/10.1007/s43681-024-00420-x

\bibitem{b5} Kang, D., Raghavan, D., Bailis, P., \& Zaharia, M. (2020). Model Assertions for Monitoring and Improving ML Models. https://doi.org/10.48550/ARXIV.2003.01668

\bibitem{b6} Paleyes, A., Urma, R.-G., \& Lawrence, N. D. (2020). Challenges in Deploying Machine Learning: A Survey of Case Studies. https://doi.org/10.48550/ARXIV.2011.09926

\bibitem{b7} Hendrycks, D. (2024). Introduction to AI Safety, Ethics, and Society (Version 2). arXiv. https://doi.org/10.48550/ARXIV.2411.01042

\bibitem{b8} Ogrizović, M., Drašković, D., \& Bojić, D. (2024). Quality assurance strategies for machine learning applications in big data analytics: An overview. Journal of Big Data, 11(1), 156. https://doi.org/10.1186/s40537-024-01028-y

\bibitem{b9} Ferreira, Raul Sena. (2023). Runtime safety monitoring of ML-based perception functions in autonomous systems [Doctoral dissertation, Université Paul Sabatier-Toulouse III]. https://theses.hal.science/tel-04342588/

\bibitem{b10} Naveed, H., Grundy, J., Arora, C., Khalajzadeh, H., \& Haggag, O. (2024). Towards Runtime Monitoring for Responsible Machine Learning using Model-driven Engineering. Proceedings of the ACM/IEEE 27th International Conference on Model Driven Engineering Languages and Systems, 195–202. https://doi.org/10.1145/3640310.3674092

\bibitem{b11} Xu, Z., Wang, R., Balaji, G., Bundele, M., Liu, X., Liu, L., \& Wang, T. (2023). AlerTiger: Deep Learning for AI Model Health Monitoring at LinkedIn. Proceedings of the 29th ACM SIGKDD Conference on Knowledge Discovery and Data Mining, 5350–5359. https://doi.org/10.1145/3580305.3599802

\bibitem{b12} Ginart, T., Jinye Zhang, M., \& Zou, J. (2022). MLDemon:deployment monitoring for machine learning systems. In G. Camps-Valls, F. J. R. Ruiz, \& I. Valera (Eds.), Proceedings of the 25th international conference on artificial intelligence and statistics (Vol. 151, pp. 3962–3997). PMLR. https://proceedings.mlr.press/v151/ginart22a.html

\bibitem{b13} Chan, A., Salganik, R., Markelius, A., Pang, C., Rajkumar, N., Krasheninnikov, D., Langosco, L., He, Z., Duan, Y., Carroll, M., Lin, M., Mayhew, A., Collins, K., Molamohammadi, M., Burden, J., Zhao, W., Rismani, S., Voudouris, K., Bhatt, U., … Maharaj, T. (2023). Harms from Increasingly Agentic Algorithmic Systems. https://doi.org/10.48550/ARXIV.2302.10329

\bibitem{b14} Rausand, M., Barros, A., \& Høyland, A. (2021). System reliability theory: Models, statistical methods, and applications (Third edition). Wiley.

\bibitem{b15} Burnell, R., Schellaert, W., Burden, J., Ullman, T. D., Martinez-Plumed, F., Tenenbaum, J. B., Rutar, D., Cheke, L. G., Sohl-Dickstein, J., Mitchell, M., Kiela, D., Shanahan, M., Voorhees, E. M., Cohn, A. G., Leibo, J. Z., \& Hernandez-Orallo, J. (2023). Rethink reporting of evaluation results in AI. Science, 380(6641), 136–138. https://doi.org/10.1126/science.adf6369

\bibitem{b16} Acharya, D. B., Kuppan, K., \& Divya, B. (2025). Agentic AI: Autonomous Intelligence for Complex Goals—A Comprehensive Survey. IEEE Access, 13, 18912–18936. https://doi.org/10.1109/ACCESS.2025.3532853

\bibitem{b17} Khamis, A. (2025). Agentic AI Systems: Architecture and Evaluation Using a Frictionless Parking Scenario. IEEE Access, 13, 126052–126069. https://doi.org/10.1109/ACCESS.2025.3590264

\bibitem{b18} Hosseini, S., \& Seilani, H. (2025). The role of agentic AI in shaping a smart future: A systematic review. Array, 26, 100399. https://doi.org/10.1016/j.array.2025.100399

\bibitem{b19} Olujimi, P. A., Owolawi, P. A., Mogase, R. C., \& Wyk, E. V. (2025). Agentic AI Frameworks in SMMEs: A Systematic Literature Review of Ecosystemic Interconnected Agents. AI, 6(6), 123. https://doi.org/10.3390/ai6060123

\bibitem{b20} Murugesan, S. (2025). The Rise of Agentic AI: Implications, Concerns, and the Path Forward. IEEE Intelligent Systems, 40(2), 8–14. https://doi.org/10.1109/MIS.2025.3544940

\bibitem{b21} Klaise, J., Van Looveren, A., Cox, C., Vacanti, G., \& Coca, A. (2020). Monitoring and explainability of models in production (Version 1). arXiv. https://doi.org/10.48550/ARXIV.2007.06299

\bibitem{b22} Heyn, H.-M., Knauss, E., Malleswaran, I., \& Dinakaran, S. (2023). An Investigation of Challenges Encountered When Specifying Training Data and Runtime Monitors for Safety Critical ML Applications. In A. Ferrari \& B. Penzenstadler (Eds.), Requirements Engineering: Foundation for Software Quality (Vol. 13975, pp. 206–222). Springer Nature Switzerland. https://doi.org/10.1007/978-3-031-29786-1\_14

\bibitem{b23} Bengio, Y., Mindermann, S., Privitera, D., Besiroglu, T., Bommasani, R., Casper, S., Choi, Y., Fox, P., Garfinkel, B., Goldfarb, D., Heidari, H., Ho, A., Kapoor, S., Khalatbari, L., Longpre, S., Manning, S., Mavroudis, V., Mazeika, M., Michael, J., … Zeng, Y. (2025). International AI Safety Report (No. arXiv:2501.17805). arXiv. https://doi.org/10.48550/arXiv.2501.17805

\bibitem{b24} Hendrycks, D., Carlini, N., Schulman, J., \& Steinhardt, J. (2021). Unsolved Problems in ML Safety (Version 5). arXiv. https://doi.org/10.48550/ARXIV.2109.13916

\bibitem{b25} Flehmig, N., Lundteigen, M. A., \& Yin, S. (2024). Implementing Artificial Intelligence in Safety-Critical Systems during Operation: Challenges and Extended Framework for a Quality Assurance Process. IECON 2024 - 50th Annual Conference of the IEEE Industrial Electronics Society, 1–8. https://doi.org/10.1109/IECON55916.2024.10906021

\bibitem{b26} Yang, J., Zhou, K., Li, Y., \& Liu, Z. (2024). Generalized Out-of-Distribution Detection: A Survey. International Journal of Computer Vision, 132(12), 5635–5662. https://doi.org/10.1007/s11263-024-02117-4

\bibitem{b27} Lu, S., Wang, Y., Sheng, L., He, L., Zheng, A., \& Liang, J. (2024). Out-of-Distribution Detection: A Task-Oriented Survey of Recent Advances (Version 4). arXiv. https://doi.org/10.48550/ARXIV.2409.11884

\bibitem{b28} Fang, Z., Li, Y., Lu, J., Dong, J., Han, B., \& Liu, F. (2022). Is Out-of-Distribution Detection Learnable? (Version 3). arXiv. https://doi.org/10.48550/ARXIV.2210.14707

\bibitem{b29} Yang, J., Zhou, K., \& Liu, Z. (2022). Full-Spectrum Out-of-Distribution Detection (Version 1). arXiv. https://doi.org/10.48550/ARXIV.2204.05306

\bibitem{b30} Longo, L., Brcic, M., Cabitza, F., Choi, J., Confalonieri, R., Ser, J. D., Guidotti, R., Hayashi, Y., Herrera, F., Holzinger, A., Jiang, R., Khosravi, H., Lecue, F., Malgieri, G., Páez, A., Samek, W., Schneider, J., Speith, T., \& Stumpf, S. (2024). Explainable Artificial Intelligence (XAI) 2.0: A manifesto of open challenges and interdisciplinary research directions. Information Fusion, 106, 102301. https://doi.org/10.1016/j.inffus.2024.102301

\bibitem{b31} Miller, T., Howe, P., \& Sonenberg, L. (2017). Explainable AI: Beware of Inmates Running the Asylum Or: How I Learnt to Stop Worrying and Love the Social and Behavioural Sciences (Version 2). arXiv. https://doi.org/10.48550/ARXIV.1712.00547

\bibitem{b32} Nisa, U., Shirazi, M., Saip, M. A., \& Pozi, M. S. M. (2025). Agentic AI: The age of reasoning—A review. Journal of Automation and Intelligence, S2949855425000516. https://doi.org/10.1016/j.jai.2025.08.003

\bibitem{b33} Lu, J., Liu, A., Dong, F., Gu, F., Gama, J., \& Zhang, G. (2018). Learning under Concept Drift: A Review. IEEE Transactions on Knowledge and Data Engineering, 1–1. https://doi.org/10.1109/TKDE.2018.2876857

\bibitem{b34} Suárez-Cetrulo, A. L., Quintana, D., \& Cervantes, A. (2023). A survey on machine learning for recurring concept drifting data streams. Expert Systems with Applications, 213, 118934. https://doi.org/10.1016/j.eswa.2022.118934

\bibitem{b35} Goodfellow, I. J., Shlens, J., \& Szegedy, C. (2014). Explaining and Harnessing Adversarial Examples (Version 3). arXiv. https://doi.org/10.48550/ARXIV.1412.6572

\bibitem{b36} Quiñonero-Candela, J. (Ed.). (2009). Dataset shift in machine learning. MIT Press.

\bibitem{b37} Lazaridou, A., Kuncoro, A., Gribovskaya, E., Agrawal, D., Liska, A., Terzi, T., Gimenez, M., d’Autume, C. de M., Kocisky, T., Ruder, S., Yogatama, D., Cao, K., Young, S., \& Blunsom, P. (2021). Mind the Gap: Assessing Temporal Generalization in Neural Language Models (No. arXiv:2102.01951). arXiv. http://arxiv.org/abs/2102.01951

\bibitem{b38} Liu, Z., Zhou, J. P., Wang, Y., \& Weinberger, K. Q. (2023). Unsupervised Out-of-Distribution Detection with Diffusion Inpainting (No. arXiv:2302.10326). arXiv. https://doi.org/10.48550/arXiv.2302.10326

\bibitem{b39} Liu, Z., Nian, Y., Zou, H. P., Li, L., Hu, X., \& Zhao, Y. (2024). COOD: Concept-based Zero-shot OOD Detection (Version 1). arXiv. https://doi.org/10.48550/ARXIV.2411.13578

\bibitem{b40} Radford, A., Kim, J. W., Hallacy, C., Ramesh, A., Goh, G., Agarwal, S., Sastry, G., Askell, A., Mishkin, P., Clark, J., Krueger, G., \& Sutskever, I. (2021). Learning Transferable Visual Models From Natural Language Supervision (Version 1). arXiv. https://doi.org/10.48550/ARXIV.2103.00020

\bibitem{b41} Jiang, X., Liu, F., Fang, Z., Chen, H., Liu, T., Zheng, F., \& Han, B. (2024). Negative Label Guided OOD Detection with Pretrained Vision-Language Models (Version 1). arXiv. https://doi.org/10.48550/ARXIV.2403.20078

\bibitem{b42} DeepSeek-AI, Liu, A., Feng, B., Xue, B., Wang, B., Wu, B., Lu, C., Zhao, C., Deng, C., Zhang, C., Ruan, C., Dai, D., Guo, D., Yang, D., Chen, D., Ji, D., Li, E., Lin, F., Dai, F., … Pan, Z. (2025). DeepSeek-V3 Technical Report (No. arXiv:2412.19437). arXiv. https://doi.org/10.48550/arXiv.2412.19437

\bibitem{b43} Miller, T. (2019). Explanation in artificial intelligence: Insights from the social sciences. Artificial Intelligence, 267, 1–38. https://doi.org/10.1016/j.artint.2018.07.007

\bibitem{b44} Simonyan, K., Vedaldi, A., \& Zisserman, A. (2013). Deep Inside Convolutional Networks: Visualising Image Classification Models and Saliency Maps (Version 2). arXiv. https://doi.org/10.48550/ARXIV.1312.6034

\bibitem{b45} Adebayo, J., Gilmer, J., Muelly, M., Goodfellow, I., Hardt, M., \& Kim, B. (2018). Sanity Checks for Saliency Maps (Version 3). arXiv. https://doi.org/10.48550/ARXIV.1810.03292

\bibitem{b46} Wei, J., Wang, X., Schuurmans, D., Bosma, M., Ichter, B., Xia, F., Chi, E., Le, Q., \& Zhou, D. (2022). Chain-of-Thought Prompting Elicits Reasoning in Large Language Models (Version 6). arXiv. https://doi.org/10.48550/ARXIV.2201.11903

\bibitem{b47} Shojaee, P., Mirzadeh, I., Alizadeh, K., Horton, M., Bengio, S., \& Farajtabar, M. (2025). The Illusion of Thinking: Understanding the Strengths and Limitations of Reasoning Models via the Lens of Problem Complexity (Version 1). arXiv. https://doi.org/10.48550/ARXIV.2506.06941

\bibitem{b48} Turpin, M., Michael, J., Perez, E., \& Bowman, S. R. (2023). Language Models Don’t Always Say What They Think: Unfaithful Explanations in Chain-of-Thought Prompting (Version 2). arXiv. https://doi.org/10.48550/ARXIV.2305.04388

\bibitem{b49} Wang, K., Variengien, A., Conmy, A., Shlegeris, B., \& Steinhardt, J. (2022). Interpretability in the Wild: A Circuit for Indirect Object Identification in GPT-2 small (Version 1). arXiv. https://doi.org/10.48550/ARXIV.2211.00593

\bibitem{b50} Olah, C., Cammarata, N., Schubert, L., Goh, G., Petrov, M., \& Carter, S. (2020). Zoom In: An Introduction to Circuits. Distill, 5(3), 10.23915/distill.00024.001. https://doi.org/10.23915/distill.00024.001

\bibitem{b54} Meng, K., Bau, D., Andonian, A., \& Belinkov, Y. (2022). Locating and Editing Factual Associations in GPT (Version 5). arXiv. https://doi.org/10.48550/ARXIV.2202.05262

\bibitem{b51} Heap, T., Lawson, T., Farnik, L., \& Aitchison, L. (2025). Sparse Autoencoders Can Interpret Randomly Initialized Transformers (No. arXiv:2501.17727). arXiv. https://doi.org/10.48550/arXiv.2501.17727

\bibitem{b52} Paulo, G., \& Belrose, N. (2025). Sparse Autoencoders Trained on the Same Data Learn Different Features (No. arXiv:2501.16615). arXiv. https://doi.org/10.48550/arXiv.2501.16615

\bibitem{b53} Zou, A., Phan, L., Chen, S., Campbell, J., Guo, P., Ren, R., Pan, A., Yin, X., Mazeika, M., Dombrowski, A.-K., Goel, S., Li, N., Byun, M. J., Wang, Z., Mallen, A., Basart, S., Koyejo, S., Song, D., Fredrikson, M., … Hendrycks, D. (2023). Representation Engineering: A Top-Down Approach to AI Transparency (Version 4). arXiv. https://doi.org/10.48550/ARXIV.2310.01405

\end{thebibliography}
\end{document}